# Can Vision-Language Models Judge Olympic Diving? From Reasoning to Scores in Zero-Shot Action Quality Assessment

Henry O. Velesaca[1,2][a], David Freire-Obregón[3][b], Luigi Miranda[1][c] and Abel Reyes-Angulo[4][d]

[1]*ESPOL Polytechnic University, Campus Gustavo Galindo, Guayaquil, Ecuador*

[2]*Software Engineering Department, Research Center for Information and Communication Technologies (CITIC-UGR), University of Granada, 18071, Granada, Spain*

[3]*SIANI, Universidad de Las Palmas de Gran Canaria, Spain*

[4]*Michigan Technological University, Houghton MI*

*hvelesac@espol.edu.ec, david.freire@ulpgc.es, luidamir@espol.edu.ec, areyesan@mtu.edu*

Keywords: Olympic sports, Judging systems, Sports analytics, Athletic performance evaluation, Video understanding.

Abstract: Automated action quality assessment (AQA) in Olympic sports remains a challenging task due to the complexity of human motion and the subjectivity inherent in expert judging. This work evaluates the capability of open-source Vision-Language Models (VLMs) to perform zero-shot action quality assessment on Olympic diving videos using the AQA-7 benchmark dataset. In this regard, a regression-based framework is proposed to leverage both the semantic reasoning and phase-level sub-scores generated by the VLMs, combining TF-IDF vectorization, dimensionality reduction, and ensemble learning to predict final competition scores. Experimental results show that standalone VLMs achieve moderate Spearman correlations below 0.32, while the proposed ensemble regression framework substantially improves performance in the reported evaluation, reaching a Spearman correlation of 0.67 with a four-model configuration. Textual reasoning features consistently outperformed raw numerical sub-scores, highlighting the richness of VLM-generated explanations for action quality analysis. These findings suggest that VLMs hold strong potential as assistive tools for explainable and semi-automated sports performance evaluation. The code is publicly available on GitHub https://github.com/hvelesaca/olympic_diving_judge_vlm

## 1 INTRODUCTION

The Olympic Games are among the most influential sporting events worldwide, attracting billions of viewers and generating substantial economic impact through broadcasting rights, sponsorships, licensing agreements, and ticket sales. According to the International Olympic Committee (IOC), the Olympic Movement generated approximately USD 7.6 billion during the 2017–2020 Olympiad cycle, with nearly 90% of these revenues redistributed to support athletes, sporting organizations, and event development worldwide[1]. More recently, the Paris 2024 Olympic Games reached an estimated audience of five billion people, corresponding to approximately 84% of the global population with access to media coverage (International Olympic Committee, 2024). Given this extraordinary visibility, the fairness and transparency of judging processes become critical, particularly in sports where performance evaluation relies heavily on subjective assessment. Disciplines such as diving and artistic gymnastics have historically faced controversies related to judging consistency and score interpretation. For example, the women's floor exercise final at the Paris 2024 Olympic Games triggered international disputes and legal appeals concerning score revisions and ranking decisions, highlighting the ongoing challenges associated with human-centered evaluation systems (Sterling, 2024).

The automatic evaluation of athletic performance has traditionally been studied under the framework of Action Quality Assessment (AQA), which aims to quantify not only the execution of an action but also its quality. Early approaches relied on hand-crafted descriptors and pose-based representations, whereas modern methods leverage deep neural net-

[a] https://orcid.org/0000-0003-0266-2465
[b] https://orcid.org/0000-0003-2378-4277
[c] https://orcid.org/0009-0003-9227-3590
[d] https://orcid.org/0000-0003-0332-8231

[1] https://olympics.com/ioc/funding

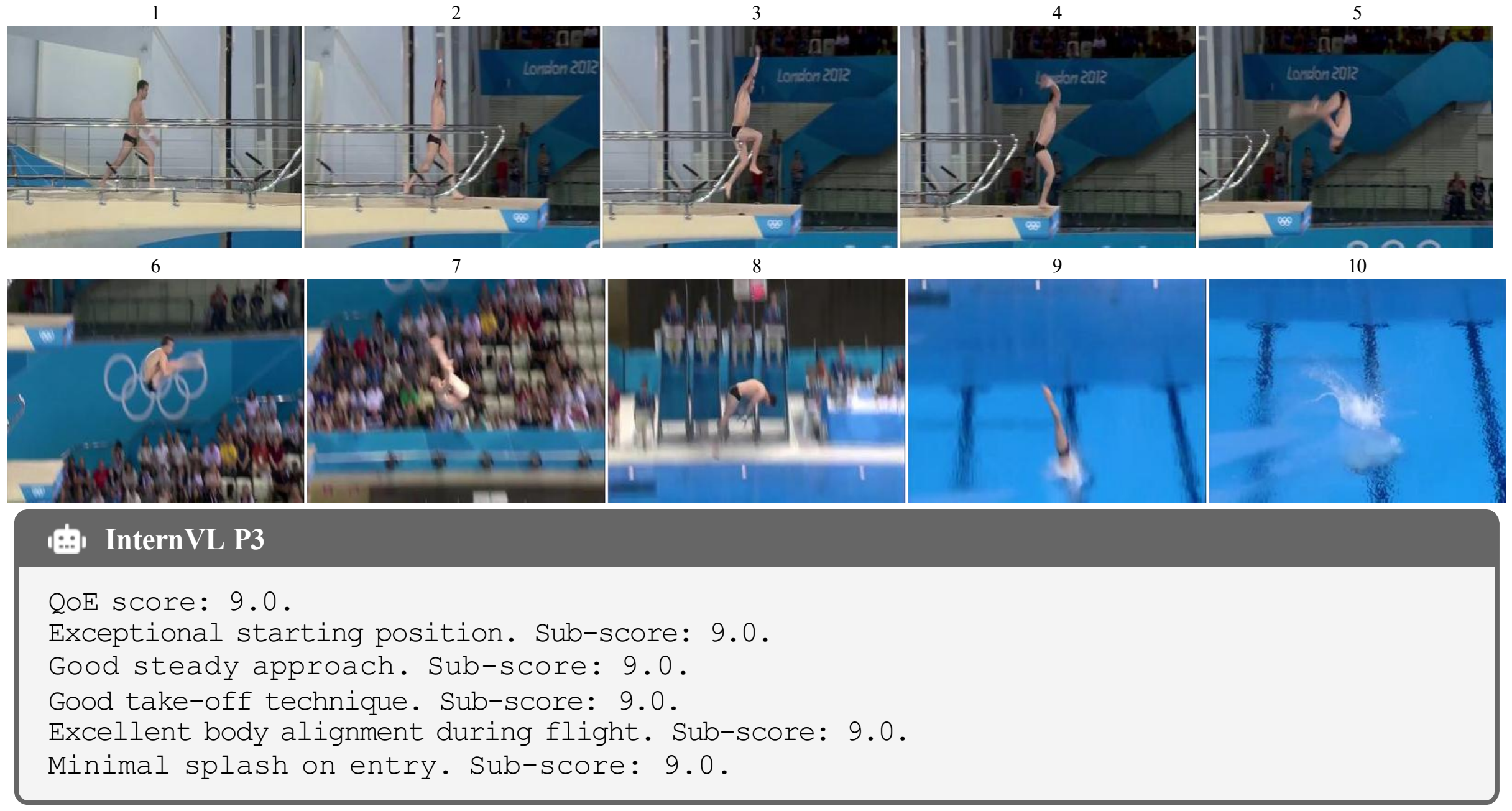

**InternVL P3**

```
QoE score: 9.0.
Exceptional starting position. Sub-score: 9.0.
Good steady approach. Sub-score: 9.0.
Good take-off technique. Sub-score: 9.0.
Excellent body alignment during flight. Sub-score: 9.0.
Minimal splash on entry. Sub-score: 9.0.
```

**Qwen 3B P2**

```
QoE score: 8.4.
The diver’s starting position is straight and aligned with the diving board, indicating
proper form. Sub-score: 8.0.
The diver’s approach was steady and consistent, showing good technique. Sub-score: 8.0.
The take-off was executed smoothly with one leg extended forward, demonstrating control and
timing. Sub-score: 9.0.
The flight phase showed a smooth and controlled rotation, indicating excellent form and
execution. Sub-score: 8.0.
The entry was minimal, with a small splash, indicating a well-executed dive. Sub-score:
8.0.
```

**LLaVA P1**

```
QoE score: 8.0.
The diver started with good posture; arms fixed and legs together; and the entry was stable
and controlled. Sub-score: 8.0.
The diver’s rhythm was steady; and the approach was consistent. Sub-score: 8.0.
The diver’s body was well-balanced; and the take-off was clean. Sub-score: 8.0.
The diver’s body was well-aligned; and the flight was controlled. Sub-score: 8.0.
The diver entered the water with control and a good shape. Sub-score: 8.0.
```

**InternVL P2**

```
QoE score: 9.5.
The diver starts correctly with no defects, earning a perfect score. Sub-score: 10.0.
The diver’s approach is steady and aligned, earning a perfect score. Sub-score: 10.0.
The take-off is clean and fully extended, earning a perfect score. Sub-score: 10.0.
The diver maintains a streamlined position with minimal rotation, earning a slight deduction.
Sub-score: 9.5.
The entry is precise with minimal splash, earning a slight deduction. Sub-score: 9.5.
```

Figure 1: Example of phase-level reasoning generated by different VLMs for a diving sequence. The generated assessments and sub-scores are subsequently used as inputs to our regression framework for AQA.

works, temporal transformers, graph-based architectures, and contrastive learning techniques to estimate performance scores directly from video sequences (Wang et al., 2022; Yin et al., 2025a). In judged Olympic sports such as diving, however, this task remains particularly challenging. Most competitions are recorded under similar environmental conditions, while decisive performance differences arise from subtle variations in body posture, rotational control, entry angle, splash generation, and temporal coordination. To address these challenges, FineDiving introduced fine-grained semantic and temporal annotations that enable procedure-aware assessment of diving performances (Xu et al., 2022b). More recently, FineParser demonstrated that accurate AQA requires detailed spatio-temporal parsing of athlete actions rather than relying solely on global video representations (Xu et al., 2024a). Despite these advances, existing supervised approaches remain fundamentally dependent on annotated datasets and historical judging scores. Consequently, they may inherit biases present in training data and often struggle to generalize across competitions, camera setups, athlete populations, or evolving judging criteria.

These limitations motivate exploring Vision-Language Models (VLMs) as an alternative paradigm for athletic performance assessment. Unlike conventional supervised AQA methods, a zero-shot VLM can be prompted with official judging guidelines and asked to reason about visual evidence without being explicitly trained to reproduce historical scores. This approach offers several potential advantages. First, it reduces dependence on large-scale labeled datasets, whose construction requires extensive expert annotation. Second, by avoiding direct optimization on historical scores, it may mitigate dataset-specific biases and provide assessments that are more closely aligned with formal evaluation criteria. Third, VLMs naturally generate textual explanations, enabling transparent analysis of execution quality, including take-off stability, body alignment, rotational precision, twist completion, and water entry characteristics. Recent advances in multimodal foundation models have demonstrated promising zero-shot visual reasoning capabilities across a wide range of domains (Wu et al., 2023; Xia et al., 2025). Moreover, recent research in sports video understanding has shown the effectiveness of athlete-centered representations, such as foreground instance selection and depth estimation, for capturing fine-grained movement characteristics without relying exclusively on pose estimation pipelines (Finocchiaro et al., 2025).

As illustrated in Figure 1, this work investigates two related settings for the assessment of Olympic diving performances. First, we evaluate the direct zero-shot scoring capability of frozen VLMs, which generate structured phase-level assessments without task-specific fine-tuning. Second, we analyze whether these zero-shot textual assessments and numerical sub-scores can be transformed into more reliable predictions through a lightweight supervised calibration model trained on the labeled AQA-7 training split. This distinction allows us to evaluate both the standalone judging capability of current VLMs and the usefulness of their generated reasoning as a representation for downstream AQA. Accordingly, throughout the paper, "zero-shot" refers to the VLM inference stage, whereas results involving trained regressors are reported as supervised calibration results.

The manuscript is organized as follows. Section 2 reviews related work on action quality assessment in sports, Vision-Language Models, and prompt engineering strategies for video understanding. Section 3 describes the proposed framework, including the dataset, selected VLM architectures, prompt design, feature extraction pipeline, and regression-based score prediction approach. Experimental results and comparisons across model configurations, ensemble strategies, and prompting variants are presented in Section 5. Finally, Section 6 summarizes the main findings and outlines directions for future research.

## 2 BACKGROUND

This section provides the necessary background for the proposed approach. We first review the development of AQA methods in sports, followed by recent advances in VLMs for video understanding and the role of prompt engineering in enabling structured multimodal reasoning.

### 2.1 AQA in Sports

AQA aims to automatically evaluate not only which action is being performed, but also how well it is executed. Unlike conventional action recognition, which focuses on categorizing activities, AQA seeks to quantify execution quality by considering both the difficulty of the action and the excellence of its execution. In sports such as diving, gymnastics, figure skating, and ski jumping, this task is particularly challenging because performance differences often arise from subtle variations in body posture, temporal coordination, landing precision, and overall movement control. As a result, AQA has emerged as a dedicated research field within computer vision and video understanding over the last decade (Yin et al., 2025b).

Early AQA research is dominated by supervised learning approaches trained to regress judging scores directly from video features. One of the pioneering works introduced the first large-scale framework for assessing Olympic diving quality using visual representations extracted from video sequences (Pirsiavash et al., 2014). Subsequent studies adopted convolutional neural networks (CNNs), recurrent neural networks (RNNs), temporal modeling architectures, and ranking-based objectives to learn mappings between athlete performances and expert-provided scores.

A major breakthrough in this direction is achieved by Parmar and Morris (Parmar and Tran Morris, 2017), who introduced one of the first deep-learning-based frameworks for sports scoring. Their approach leveraged spatio-temporal representations extracted from 3D convolutional networks and recurrent architectures to assess diving, gymnastics vault, and figure skating performances, significantly outperforming previous pose-based methods and demonstrating the effectiveness of deep neural networks for AQA.

As the field matured, researchers recognized that global video representations are often insufficient for capturing the fine-grained details that influence judging decisions in Olympic sports. This limitation motivated the development of specialized datasets and architectures designed to model action phases and procedural structure. FineDiving introduced temporally annotated diving sequences together with fine-grained semantic labels, enabling procedure-aware assessment of diving performances (Xu et al., 2022a). Building upon this idea, FineParser proposed a spatio-temporal action parsing framework capable of explicitly modeling the different stages of athletic movements, demonstrating that accurate AQA requires detailed understanding of localized motion patterns rather than relying solely on holistic video features (Xu et al., 2024b).

Despite the considerable progress achieved by supervised approaches, several limitations remain. Most state-of-the-art methods depend on large annotated datasets and historical judging scores, making them costly to develop and potentially susceptible to inheriting biases present in training annotations. Furthermore, their ability to generalize across sports, competitions, camera viewpoints, athlete populations, and evolving judging criteria remains limited. Recent surveys identify explainability, generalization, fine-grained reasoning, and reduced dependence on annotated data as key open challenges in AQA research (Yin et al., 2025b). These limitations motivate the exploration of VLMs, which can leverage multimodal reasoning capabilities to analyze athletic performances in a zero-shot setting and provide natural-language explanations that are more transparent and interpretable than conventional score-regression approaches.

## 2.2 Vision-Language Models for Video Understanding

VLMs are multimodal artificial intelligence models that combine visual input (such as images and videos) with textual input. This integration allows machines to perceive and reason about the world through both modalities, providing a more contextual understanding of spatial relationships, objects, scenes, and abstract concepts compared to single-modal models. Thanks to this capability, VLMs can be applied to multimodal action understanding and quality-assessment tasks. (Li et al., 2025)

The evolution of VLMs from the processing of static images to the understanding and generating dynamic videos is driven by several key advancements. The foundational vision encoders have helped these models extract visual features from both image and video data, enabling them to effectively integrate continuous, temporal sequences with textual language representations. For the VLMs in order to comprehend long videos, the models require to capture fine-grained spatiotemporal details, reason between different events, and track long-term dependencies over extended periods. (Zou et al., 2024) The evolution of VLMs can be traced through the development of progressively more complex architectures, highlighting key models such as LLaVA, Qwen-VL, MiniCPM-V, and Video-LLaVA.

LLaVA (Large Language and Vision Assistant) operates as a VLM by connecting a pre-trained vision encoder with a Large Language Model (LLM) to achieve general-purpose visual and language understanding(Liu et al., 2023), it utilizes the pre-trained CLIP visual encoder to extract visual features from input images. To bridge the gap between the visual and textual modalities, it employs a simple, lightweight linear projection layer. LLaVA represents the first attempt to extend the concept of ”instruction tuning” a technique that drastically improved the zero-shot capabilities of text-only NLP models—into the language-image multi-modal space. The LLaVA-Next, newer version, expands the capabilities of open-source models to act as a generalist evaluator. Instead of just answering questions about an image, it can evaluate the quality of responses generated by other multimodal models. It provides reliable quantitative scores and model rankings. (Zhang et al., 2024)

Qwen2.5-VL is designed to act as an advanced, interactive visual agent capable of understanding and reasoning about both static and dynamic environments. It is available in three sizes (3B, 7B, and 72B parameters). This model excels at spatial reasoning and can accurately detect, point to, and count objects using bounding boxes or precise point coordinates based on the actual dimensions of the image. (Bai et al., 2025)

On the other hand, MiniCPM-V is a series of highly efficient Multimodal Large Language Models (MLLMs) explicitly designed to run locally on devices, such as mobile phones and personal computers, rather than relying on massive cloud servers. To make running on a phone possible, MiniCPM-V employs a shared compression layer that shrinks the visual tokens of each image slice down to just 64 to 96 tokens. (Yao et al., 2024)

Most conventional MMLMs first train an LLM and later retrofit it to understand visual inputs, which often causes alignment issues and compromises the core language abilities. InternVL3, instead, is jointly trained on both large-scale text-only corpora and diverse multimodal datasets from the very beginning.By updating all model parameters together, the linguistic and visual features evolve synchronously, allowing the model to naturally handle both text and vision. (Chen et al., 2025)

# 3 METHODOLOGY PROPOSED

This section describes the proposed pipeline used to evaluate VLMs for zero-shot action quality assessment in Olympic diving, illustrated in Figure 2. The pipeline consists of a sequence of stages, including VLM selection, prompt design, video inference, reasoning and score extraction, regression-based score prediction, and performance evaluation. Each stage is described in detail in the remainder of this section.

**Selected VLMs**. We evaluate six open-source vision-language models (VLMs) for video and text understanding, described in Section 2.2: Qwen2.5-VL-3B-Instruct, Qwen2.5-VL-7B-Instruct, Qwen2.5-VL-32B-Instruct, LLaVA-NeXT-Video-7B, MiniCPM-V 2.6, and InternVL3-8B. The selection criteria prioritized architectures capable of natively processing video inputs and generating detailed natural-language reasoning, both of which are essential for the feature extraction framework described in subsequent sections. The evaluated models span parameter scales from 3B to 32B and encompass several state-of-the-art VLM families with diverse visual encoders, multimodal fusion strategies, and training methodologies.

**Prompt Engineering**. A collection of structured prompts is designed to elicit phase-level assessments from VLMs analyzing competitive diving videos. Each prompt instructed the models to generate both numerical sub-scores and qualitative evaluations for the five canonical phases of a dive: starting position, approach, take-off, flight, and entry.

Prompt development followed an iterative refinement process aimed at improving output specificity, consistency, and interpretability. Successive prompt versions are evaluated based on output structure, score distributions, and semantic coherence. Key design elements included explicit scoring scales, phase-specific evaluation criteria, and constrained output formats enforced through a predefined JSON schema, illustrated in Figure 3 to facilitate automated downstream processing. The final prompt configurations considered in the experiments are three different prompts.

**VLM Execution on Video Data**. Each VLM configuration is evaluated across the complete dataset, processing one video clip per inference call. Models received both the video input and the corresponding prompt, and are instructed to perform step-by-step reasoning across the different diving phases before generating a structured assessment. Inference is conducted independently for every combination of model architecture, parameter scale, and prompt variant, producing one reasoning output per video. Average inference time is recorded for each configuration to characterize computational requirements.

**Score and Reasoning Extraction from VLM Outputs**. A rule-based extraction pipeline is developed to retrieve numerical sub-scores from the free-text outputs generated by the VLMs. Regular expressions and structured pattern-matching techniques are employed to identify phase-level scores while accommodating formatting variations across model families. Extracted values are validated against the expected scoring range; invalid or missing outputs are flagged and subsequently imputed when necessary.

In addition to numerical scores, the complete natural-language reasoning generated by each model is extracted and stored for every video and diving phase. These textual explanations constitute the primary source of semantic information used in subsequent analyses. Prior to feature extraction, reasoning texts are cleaned, normalized, and aligned with their corresponding video identifiers and ground-truth annotations.

**Regressor Training Using VLM Reasoning and Phase Sub-scores**. To predict final diving scores, a supervised regression framework is developed using two complementary feature modalities: (1) semantic representations derived from VLM-generated

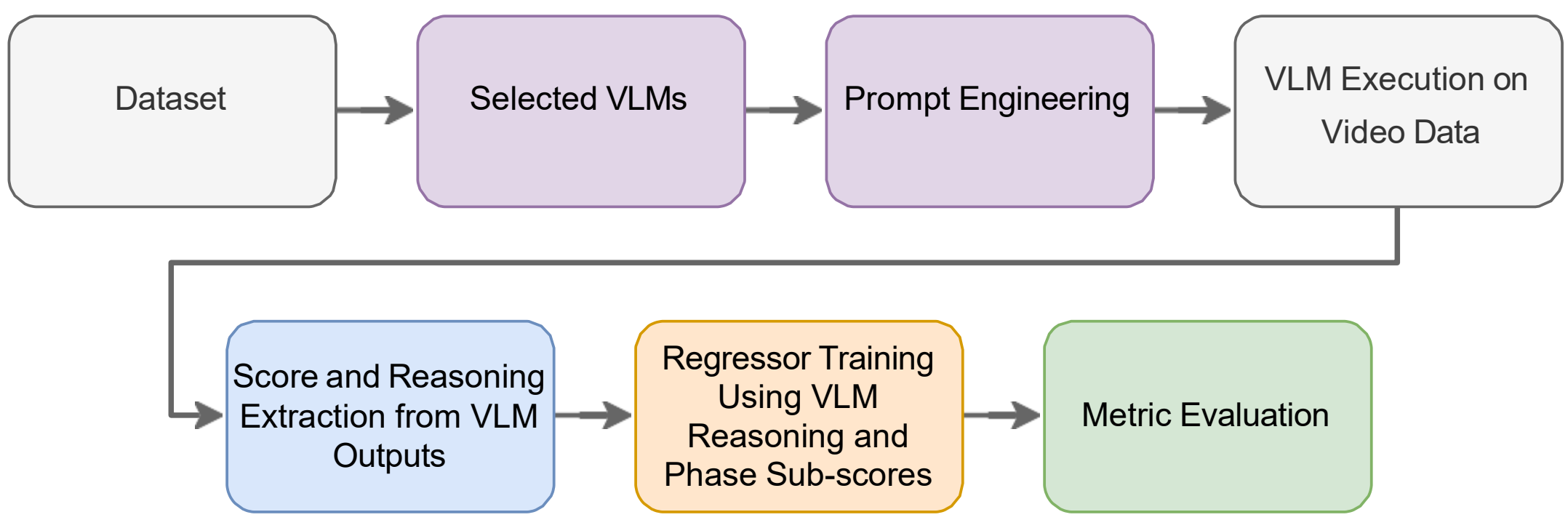

Figure 2: Overview of the two-stage framework. Frozen VLMs first generate zero-shot phase-level assessments from diving videos. Their textual reasoning and numerical sub-scores are then used as features by a supervised calibration model trained on the labeled AQA-7 training split.

```
{
  "Phase_Analysis": {
    "Starting_Position": {
      "observations": ["<fact 1>", "<fact 2>"],
      "defects_present": ["<defect>"],
      "sub_score": <float 0.0-10.0>
    },
    "Approach": {
      "observations": ["<fact 1>", "<fact 2>"],
      "defects_present": ["<defect>"],
      "sub_score": <float 0.0-10.0>
    },
    "Take_off": {
      "observations": ["<fact 1>", "<fact 2>"],
      "defects_present": ["<defect>"],
      "sub_score": <float 0.0-10.0>
    },
    "Flight": {
      "observations": ["<fact 1>", "<fact 2>"],
      "defects_present": ["<defect>"],
      "sub_score": <float 0.0-10.0>
    },
    "Entry": {
      "observations": ["<fact 1>", "<fact 2>"],
      "defects_present": ["<defect>"],
      "sub_score": <float 0.0-10.0>
    }
  },
  "QoE": <must equal final_qoe>,
  "QoE_Justification": {
    "Starting_Position": "<concise summary tied to observations + sub-score>",
    "Approach": "<concise summary + sub-score>",
    "Take_off": "<concise summary + sub-score>",
    "Flight": "<concise summary + sub-score>",
    "Entry": "<concise summary + sub-score>"
  }
}
```

Figure 3: Structured JSON schema used to obtain phase-level VLM assessments of Olympic diving performances.

reasoning texts, and (2) numerical phase-level sub-scores extracted from model outputs. Semantic features are obtained through TF-IDF vectorization followed by dimensionality reduction via Truncated Singular Value Decomposition (SVD). Multiple regression algorithms are evaluated: Random Forest (RF), Gradient Boosting Regressor (GBR), Ridge Regression, and Support Vector Regression (SVR). Models are trained on the predefined 70% training split, and hyperparameters are optimized through five-fold cross-validation.

To systematically assess the predictive contribution of individual VLMs and their combinations, an automated experimentation framework is imple-mented in Python. The framework exhaustively explored all possible combinations of predefined VLM sources across configurable ensemble sizes, training and evaluating a regression model for each configuration.

**Data sources and preprocessing.** Reasoning outputs are collected from multiple VLM configurations spanning different model families, each evaluated under different prompting strategies. For every video, each model generated phase-level sub-scores and qualitative textual assessments describing the perceived quality of execution across the five diving phases.

To assess the contribution of model diversity, experiments are conducted across ensemble sizes from 1 to 4 (i.e., GROUP_SIZES = [1, 2, 3, 4]). For size 1, each VLM configuration is evaluated independently. For sizes 2 through 4, all ordered permutations of the available VLM sources are enumerated, yielding a total of N experiments per size. This exhaustive combinatorial design enables a direct comparison of single-model versus multi-model fusion strategies.

**Feature engineering.** For each experimental configuration, numerical phase-level scores are concatenated with structured metadata — specifically the dive DoD coefficient. Textual descriptions are transformed into numerical representations using TF-IDF vectorization (unigrams, bigrams, and trigrams; min_df = 3; linear term frequency weighting) followed by Truncated SVD with 30 components. Three feature sets are constructed: (1) semantic and numerical features combined (SVD + scores), (2) semantic features only (SVD), and (3) semantic features augmented with dive difficulty information.

**Model evaluation.** Several regressor configurations are evaluated per experiment, comprising RF, GBR, Ridge, and SVR models applied to the different feature representations. Performance is measured using Spearman's rank correlation coefficient between predicted and ground-truth scores.

# 4 EXPERIMENTAL SETUP

**Dataset**. This study utilizes the AQA-7 dataset (Parmar and Morris, 2019), a widely adopted benchmark for action quality assessment in Olympic sports. The dataset comprises seven disciplines (i.e., Single Diving 10m Platform, Gymnastic vault, Big Air Skiing, Big Air Snowboarding, Sync. Diving 3m springboard, Sync. Diving 10m Platform, and Trampoline).

In this work, we focus on the Single Diving 10m Platform discipline because it involves complex aerial movements, precise body control, and execution differences that are strongly reflected in the final scores, making it particularly challenging for action quality assessment methods. Moreover, each annotated video clip is associated with a ground-truth final score and a corresponding Degree of Difficulty (DoD) determined according to official competition regulations. A fixed train-test split of 300/70 samples is employed across all experiments. Training and testing indices of each video are obtained to official pre-established dataset split to ensure reproducibility and facilitate direct comparison among model configurations. In competitive diving, the final score is computed as the sum of the judges' scores multiplied by the dive DoD. This scoring procedure is consistently applied throughout the evaluation process to maintain alignment with official judging standards.

As illustrated in Figure 4, the Single Diving 10m Platform event presents significant temporal variation across frames, capturing critical phases such as the takeoff, aerial rotation, and water entry. This temporal diversity poses a challenge for quality assessment models, as subtle differences in body posture and trajectory across frames directly influence the assigned score.

**Metric Evaluation**. Predictive performance is assessed at two complementary levels. First, as a baseline, the raw sub-scores generated directly by each VLM configuration are evaluated without any regression post-processing. Spearman's rank correlation is computed between VLM-derived predictions and ground-truth annotations on the held-out test set, both in their original form and after applying the official scoring transformation. Additional metrics — Pearson correlation, Root Mean Squared Error (RMSE), Mean Absolute Error (MAE), and Relative Error (RelErr%) — are computed to provide a comprehensive assessment of prediction quality.

Second, the trained regression models are evaluated using the same metric suite. Results are reported for each regressor type, feature representation, and VLM combination, enabling a systematic comparison of feature modalities and ensemble configurations. The best-performing setup per ensemble size is identified based on the adjusted Spearman correlation coefficient, which reflects alignment with official competition scoring criteria.

# 5 EXPERIMENTAL RESULTS

This section presents the experimental results obtained from both standalone Vision-Language Models and the proposed regression-based framework.

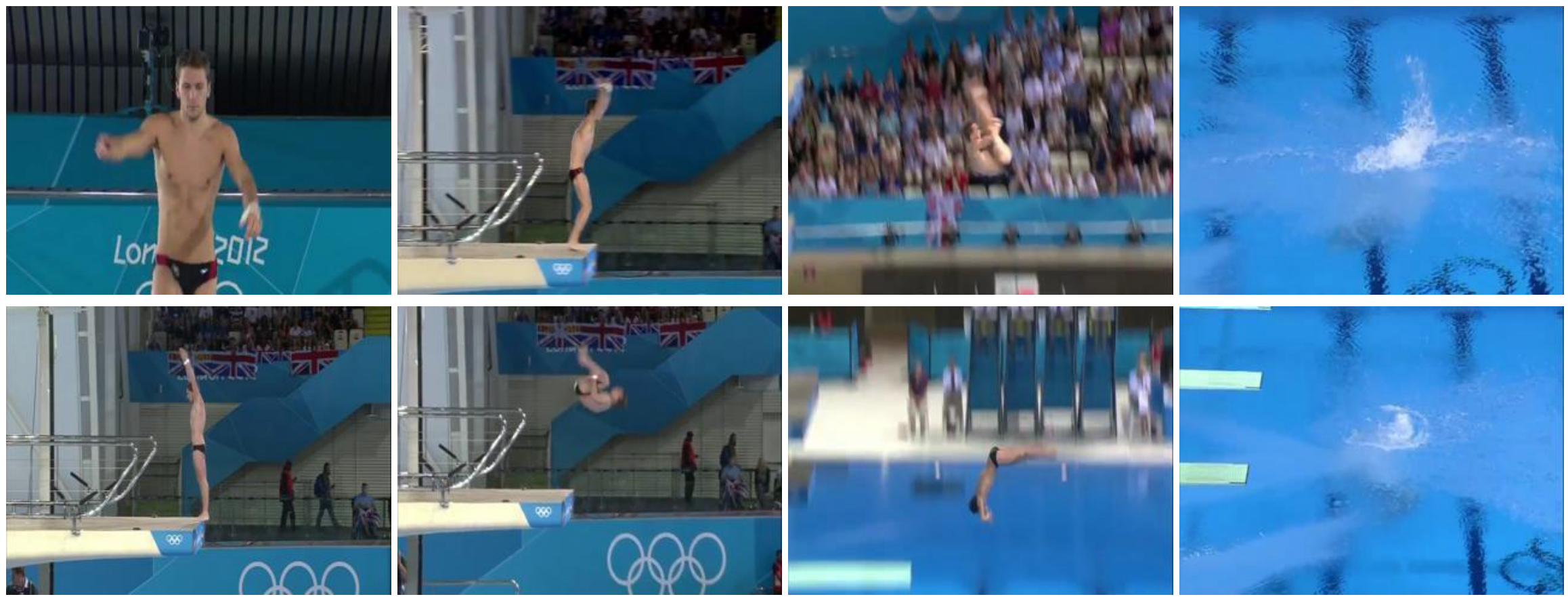

Figure 4: Sample frames extracted from the AQA-7 dataset (Parmar and Morris, 2019), illustrating the temporal progression of a Single Diving 10m Platform action sequence.

Table 1: Performance comparison of the evaluated VLM configurations and the proposed regressor ensemble on the AQA-7 Single Diving 10m Platform dataset. Higher values are better for Spearman, Pearson, and $R^2$, while lower values are better for RMSE, MAE, and Relative Error.

| Model | Spearman↑ | Pearson↑ | RMSE↓ | MAE↓ | RelErr%↓ | $R^2$↑ |
|---|---|---|---|---|---|---|
| Qwen 3B P1 | -0.1305 | -0.2668 | 30.442 | 25.457 | 37.92 | -2.8348 |
| Qwen 3B P2 | 0.2892 | 0.3481 | 15.901 | 11.374 | 19.92 | -0.0463 |
| Qwen 3B P3 | 0.2056 | -0.0093 | 20.878 | 14.692 | 23.31 | -0.8038 |
| Qwen 7B P1 | 0.2356 | 0.1249 | 17.881 | 12.571 | 22.04 | -0.3230 |
| Qwen 7B P2 | 0.2967 | 0.1068 | 19.282 | 13.212 | 24.84 | -0.5385 |
| Qwen 7B P3 | 0.2636 | 0.2328 | 29.720 | 25.374 | 33.96 | -2.6550 |
| Qwen 32B P1 | 0.3174 | 0.1924 | 22.408 | 16.008 | 29.32 | -1.0778 |
| Qwen 32B P2 | 0.2715 | 0.1751 | 20.482 | 14.755 | 26.98 | -0.7359 |
| Qwen 32B P3 | 0.2755 | 0.2253 | 18.862 | 13.574 | 24.79 | -0.4722 |
| LLaVA P1 | 0.2390 | 0.2614 | 15.455 | 11.312 | 19.63 | 0.0116 |
| LLaVA P3 | 0.2045 | 0.0757 | 31.162 | 21.733 | 31.98 | -3.0184 |
| MiniCPM P1 | 0.0288 | -0.0310 | 29.018 | 24.446 | 34.41 | -2.4844 |
| InternVL P1 | 0.2189 | 0.0984 | 17.304 | 13.494 | 21.77 | -0.2390 |
| InternVL P2 | 0.2784 | 0.1852 | 25.502 | 19.876 | 34.15 | -1.6912 |
| InternVL P3 | 0.2937 | 0.0674 | 23.782 | 16.009 | 28.12 | -1.3405 |
| Regressor Ensemble (Ours) | **0.6664** | **0.5223** | **13.564** | **9.306** | **16.99** | **0.2387** |

## 5.1 Standalone VLM Performance

The first set of experiments evaluated the raw predictions produced directly by each VLM configuration without applying any regression model. Table 1 reports the main evaluation metrics across all tested models and three prompt variants, named Prompt 1 (P1), Prompt 2 (P2), and Prompt 3 (P3).

Among the standalone configurations, Qwen 32B with P1 achieved the highest Spearman correlation (0.3174), indicating the strongest monotonic alignment with official judging scores. Qwen 3B with P2 obtained the best Pearson correlation (0.3481), while LLaVA with P1 produced the lowest RMSE (15.455) and MAE (11.312). In contrast, configurations such as Qwen 3B P1 exhibited negative correlation and large prediction error, suggesting unstable alignment with human judging criteria under those prompting conditions.

Overall, the raw VLM predictions showed only moderate correspondence with official scores, with Spearman correlations generally remaining below 0.32. This indicates that, while the models capture some aspects of dive quality, their direct outputs are insufficient for reliable score prediction without additional processing.

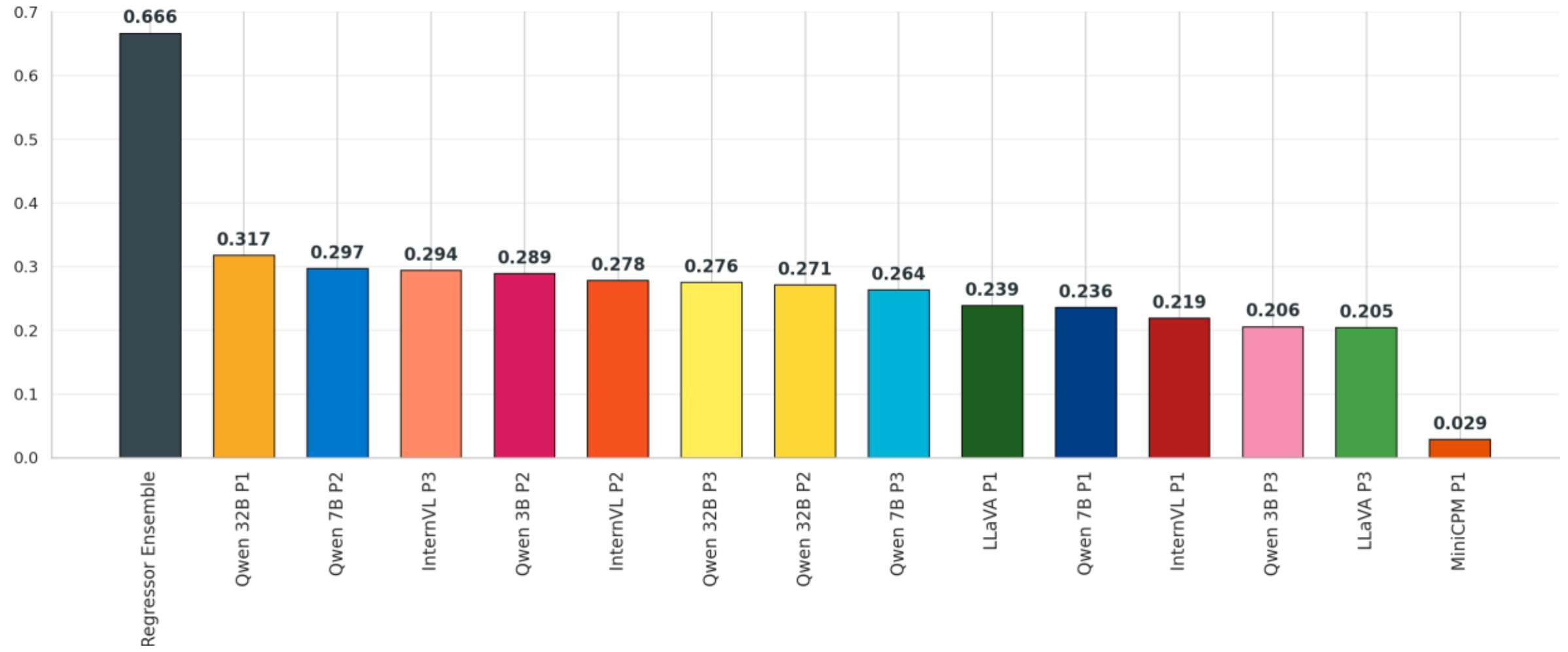

Figure 5: Spearman Pareto Chart of all VLMs.

## 5.2 Effect of Regression Models and Feature Engineering

To improve predictive performance, regression models are trained using semantic features extracted from VLM reasoning texts, numerical phase-level sub-scores, and dive difficulty coefficients. Figure 5 presents a Pareto ranking of all evaluated configurations by Spearman correlation, clearly showing that the ensemble-based regressor substantially outperforms every standalone VLM. The regression framework achieved a Spearman correlation of 0.6664, more than doubling the best standalone result (0.3174). It also attained the highest Pearson correlation (0.5223), the lowest RMSE (13.564), and the lowest MAE (9.306), demonstrating markedly closer agreement with official scores.

These results confirm that the textual reasoning generated by VLMs encodes meaningful information about dive quality, but that this information must be structured and aggregated through supervised learning to produce accurate final predictions.

## 5.3 Performance by Ensemble Size

Table 2 summarizes the top-performing VLM combinations for ensemble sizes ranging from one to four models. The results reveal a clear progressive trend: larger and more diverse ensembles consistently improved predictive accuracy. Single-model configurations reached Spearman correlations around 0.47, two-model ensembles exceeded 0.59, three-model ensembles further improved to approximately 0.61, and the best four-model ensemble achieved the overall highest correlation of 0.6664.

Another notable finding is that text-based semantic features are often more informative than numerical phase scores alone. In many of the best-performing configurations, the highest correlations are obtained using textual reasoning features without the extracted sub-scores, suggesting that the qualitative explanations generated by VLMs capture richer contextual information about dive execution than the numerical scores they directly produce.

## 5.4 Analysis of Prompting Strategies

Prompt design had a substantial impact on VLM performance. Across several model families, P2 generally yielded higher Spearman correlations than P1 or P3. P2 consistently provided a better balance between structured reasoning and scoring consistency, with lower RMSE values and stronger rank correlations. P3, by contrast, often produced less stable outputs, with higher prediction errors and weaker correlations, suggesting that overly constrained or complex prompt formulations can reduce the consistency of phase-level reasoning.

Figure 6 summarizes the complementary capabilities of the evaluated VLMs through a radar-chart comparison of six metrics: Pearson, Spearman, $R^2$ (normalized), and inverted error measures (1-RelErr, 1-MAE, 1-RMSE). The visualization confirms that no single standalone model dominates across all dimensions simultaneously, reinforcing the value of the ensemble regression approach.

## 5.5 Summary of Findings

The experimental results demonstrate that:

Table 2: Top-performing configurations for each ensemble size. For every group size, the table reports the three highest-ranking combinations of VLM sources, regression model, and input feature set, measured using Spearman's rank correlation coefficient on the AQA-7 Single Diving 10m Platform benchmark.

| Group | VLMs top-3 best combination | Regressor | Input | Spearman↑ |
|---|---|---|---|---|
| 1 | InternVL P2 | RF | txt + scores | 0.4664 |
| 1 | InternVL P2 | RF | txt | 0.4677 |
| 1 | InternVL P2 | RF | txt + scores + DoD | 0.4792 |
| 2 | Qwen 3B P2 + InternVL P2 | GBR | txt | 0.5977 |
| 2 | Qwen 3B P2 + InternVL P2 | GBR | txt + scores | 0.6022 |
| 2 | Qwen 3B P2 + InternVL P2 | GBR | all | 0.6029 |
| 3 | InternVL P1 + InternVL P2 + Qwen 3B P2 | GBR | txt + scores + DoD | 0.5988 |
| 3 | InternVL P1 + InternVL P2 + Qwen 3B P2 | GBR | txt | 0.5950 |
| 3 | InternVL P1 + InternVL P2 + Qwen 3B P2 | GBR | txt + scores | 0.6123 |
| 4 | LLaVA P1 + Qwen 7B P2 + InternVL P2 + InternVL P1 | GBR | txt | 0.6212 |
| 4 | Qwen 32B P2 + Qwen 7B P1 + LLaVA P1 + InternVL P2 | GBR | txt | 0.6397 |
| 4 | InternVL P3 + Qwen 3B P2 + LLaVA P1 + InternVL P2 | GBR | txt + scores | **0.6664** |

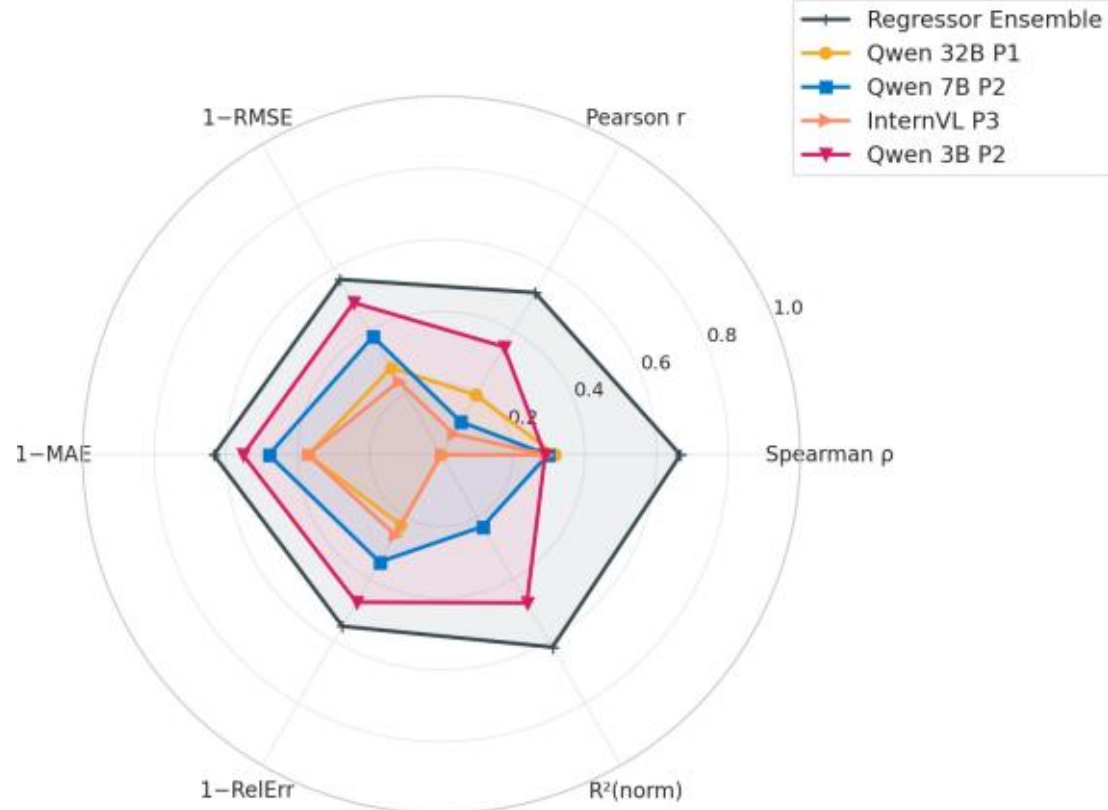

Figure 6: Radar chart comparing VLM capabilities using different metrics.

- Zero-shot VLMs can extract meaningful semantic information from Olympic diving videos, but their direct score predictions are not sufficiently reliable for standalone judging purposes.
- Regression models trained on VLM-generated reasoning and phase sub-scores significantly improve alignment with official competition scores.
- Ensemble strategies combining multiple VLM families provide the best performance, with the top four-model configuration achieving a Spearman correlation of 0.6664.
- Textual reasoning features are particularly informative, indicating that VLM explanations encode richer information about action quality than numerical sub-scores alone.
- Prompt engineering plays a critical role in stabilizing and improving VLM assessments, with P2 offering the best balance across model families.

Taken together, these findings suggest that VLMs show strong potential as assistive tools for explainable performance analysis and semi-automated action quality assessment.

# 6 CONCLUSIONS

Recent advances in multimodal foundation models suggest a shift in sports video analysis from traditional Human Action Recognition (HAR) and AQA pipelines toward approaches that combine visual perception with semantic reasoning. While conventional methods have achieved remarkable performance through supervised learning, they often rely on large annotated datasets and provide limited interpretability. In contrast, emerging approaches increasingly focus on extracting semantically meaningful representations from athletes and their actions, enabling richer descriptions of performance and more explainable decision-making processes (Finocchiaro et al., 2025). The results presented in this work indicate that VLMs can contribute to this transition by generating structured assessments that capture aspects of athletic performance beyond numerical score prediction.

The experimental evaluation reveals that, while standalone VLMs exhibit moderate ranking ability (with Spearman correlations ranging from 0.13 to 0.32), their combination through a regression-based ensemble framework significantly improves predictive performance, achieving a Spearman correlation of 0.6664 and an MAE of 9.306. Notably, the best-performing individual configuration, InternVL P2, reached a Spearman of 0.4792 when combined with text, score, and DoD features, suggesting that auxiliary structured features complement the raw VLM outputs. The consistent benefit of prompt engineering, particularly the structured sub-score P2, further demonstrates that model performance is highly sensitive to how the task is framed, highlighting the importance of prompt design in zero-shot multimodal evaluation scenarios. These findings suggest that no single model architecture or prompt strategy dominates across all metrics, and that diversity in model selection and feature representation is key to ensemble effectiveness.

Beyond Olympic judging disciplines, the ability of multimodal models to reason about human actions has potential applications across a wide range of sports analytics tasks. Recent work has shown that human-action representations can be exploited to predict player intentions and future outcomes in highly dynamic scenarios, such as early penalty direction prediction in football (Velesaca et al., 2026). This suggests that the combination of visual understanding, semantic reasoning, and temporal analysis explored in this work could extend beyond action quality assessment, supporting decision-making systems, performance analysis, and tactical understanding in mainstream sports.

## ACKNOWLEDGEMENTS

This research has been supported by the ESPOL project “Reconocimiento de patrones en ima´genes usando te´cnicas basadas en aprendizaje” (CIDIS-004-2024).